\documentclass[10pt,twocolumn,letterpaper]{article}

\usepackage{cvpr}
\usepackage{times}
\usepackage{epsfig}
\usepackage{graphicx}
\usepackage{amsmath}
\usepackage{amsfonts}
\usepackage{amssymb}
\usepackage{calligra}
\usepackage{subcaption}
\usepackage{textcomp}

\usepackage[pagebackref=true,breaklinks=true,letterpaper=true,colorlinks,bookmarks=false]{hyperref}

\cvprfinalcopy 

\def\cvprPaperID{1592} 

\ifcvprfinal\fi
\begin{document}

\title{DIY Human Action Data Set Generation}

\author{Mehran Khodabandeh \thanks{The work was performed during an internship at Microsoft.}\\
{\tt\small mkhodaba@sfu.ca}  \\
Simon Fraser University 
\and
Hamid Reza Vaezi Joze \\
{\tt\small hava@microsoft.com} \\
Microsoft
\and
Illya Zharkov\\
{\tt\scriptsize  zharkov@microsoft.com} \\
Microsoft
\and
Vivek Pradeep\\
{\tt\scriptsize  vpradeep@microsoft.com} \\
Microsoft
}

\maketitle

\begin{abstract}
  
   The recent successes in applying deep learning techniques to solve standard computer vision problems has aspired researchers to propose new computer vision problems in different domains. 
   As previously established in the field, training data itself plays a significant role in the machine learning process, especially deep learning approaches which are data hungry. In order to solve each new problem and get a decent performance, a large amount of data needs to be captured which may in many cases pose logistical difficulties. Therefore, the ability to generate de novo data or expand an existing data set, however small, in order to satisfy data requirement of current networks may be invaluable. Herein, we introduce a novel way to partition an action video clip into action, subject and context. Each part is manipulated separately and reassembled with our proposed video generation technique. Furthermore, our novel human skeleton trajectory generation along with our proposed video generation technique, enables us to generate unlimited action recognition training data. These techniques enables us to generate video action clips from an small set without costly and time-consuming data acquisition. Lastly, we prove through extensive set of experiments on two small human action recognition data sets, that this new data generation technique can improve the performance of current action recognition neural nets.      
   \vspace{-4mm}
\end{abstract}


\begin{figure}[ht!]
\begin{center}
\includegraphics[width=.90\linewidth, ]{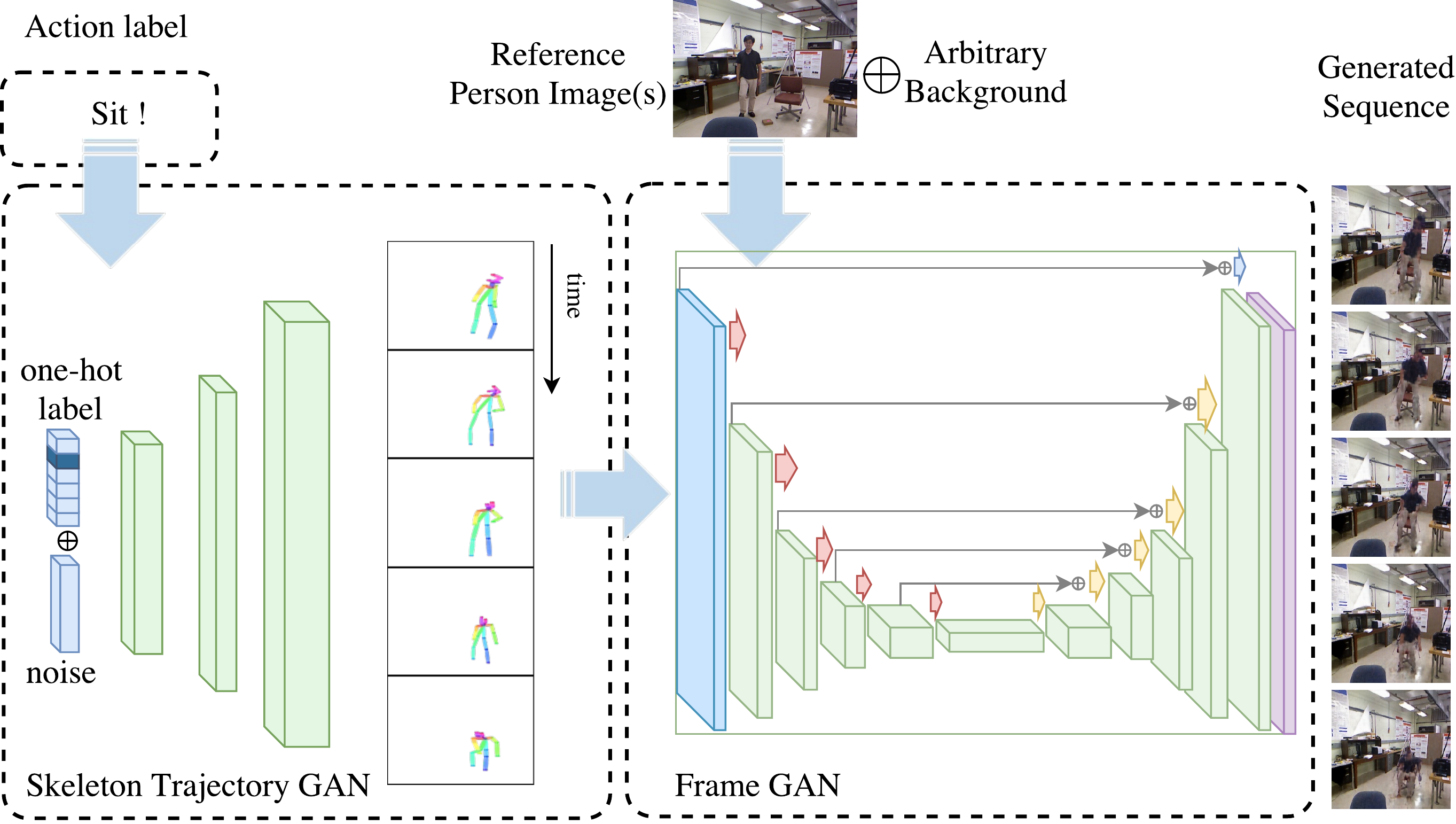}
\end{center}
   \caption{Our algorithm takes as input an action label, a set of reference images and an arbitrary background. The output is a generated video of the person in the reference image performing a given action. We approached this problem in two stages. Firstly (left side) a generative model trained on a small \textit{labeled} dataset of skeleton trajectories of human actions, generates a sequence of human skeletons conditioned on the action label. Secondly (right side), another generative mode trained on an \textit{unlabeled} set of human action videos, generates a sequence of photo-realistic frames conditioned on the given background, generated skeletons, and the person's appearance given in the reference frames. This produces an arbitrary number of human action videos.
   \vspace{8pt}
   }
\label{fig:pull_figure}
\end{figure}

\section{Introduction}
    After significant successes in face detection, face recognition and object detection commonly used in our daily life, computer vision researchers are now aiming at understanding video which is one dimension more difficult. These successes rely on advanced machine learning techniques and training data which require computational power, mainly deep networks. Hence, the process of data acquisition may be as vital as the technique used. Large data sets, such as a million object and animal photos~\cite{krizhevsky2012imagenet}, hundreds of thousands of faces~\cite{kemelmacher2016megaface} or millions of scenes~\cite{lin2014microsoft}, enables complex neural networks to train successfully. However, similar results can never be achieved through small data sets manually captured by researchers themselves. 
    Video data sets or specifically human action data sets are more difficult to compile. There are two common scenarios to generate a human action data set: (1) asking subjects to do a series of actions in front of a camera (2) labeling an existing video from the internet. The first scenario is not scaleable considering the number of subjects and the limitations imposed by the capturing environment. These types of data sets are not common anymore due to their small size. Some examples of the second scenario are UCF~101~\cite{soomro2012ucf101} containing 101 actions of thousands of online clips, Hollywood2~\cite{marszalek09} containing 12 actions in around 3 thousands clip extracted from movies and the kinetics~\cite{kay2017kinetics} including 400 actions from hundreds of thousands of YouTube videos. Although these data sets are very useful to benchmark the accuracy of different algorithms, the clips or actions are not necessarily useful for real world action recognition tasks such as security surveillance cameras, sport analysis, smart home devices, health monitoring etc, as each scenario has different settings and sets of actions. A solution would be for researchers to collect their own data sets which may prove to be costly and time consuming. 
    
    In this paper, we've introduced a novel way to partition an action video clip into action, subject and context. We showed that we can manipulate each part separately and assemble them with our proposed video generation model into new clips. The actions are represented by a series of skeletons, the context is an still image or a video clip, and the subject is represented by random images of the same person. We can change an action by extracting it from an arbitrary video clip, generate it through our proposed skeleton trajectory model, or by applying perspective transform on existing skeleton. Additionally, we can change the subject and the context using arbitrary video clips, enabling us to arbitrarily generate action clips. This is particularly useful for action recognition models which require large data sets to increase their accuracy. With the use of a large unlabeled data and a small set of labeled data, we can synthesize a realistic set of training data for training a deep model.

    We called it DIY (do it yourself) because we can eventually build our own data set from a small one. Similar to actual data collection, not only we can add a new person or action to the data set, but also internally expand the data set or capture the same data from different angles with very little time and effort.
    
   Lastly, to quantitatively evaluate our data generation technique, we applied it to UT~Kinects~\cite{xia2012view} a human action data set comprised of 10 actions in 200 video clips. We generated new video clip types by adding new subjects or actions or by expanding current action and subjects. It is shown that generated data along with the existing data, can improve the performance of well-performed video representation networks: I3D~\cite{carreira2017quo} and C3D~\cite{tran2015learning} on action recognition task. For further investigation, we applied our method and action recognition task to actions with two persons in SUB interact~\cite{yun2012two} data sets.     
   The outline of this paper is as follows. In \textsection \ref{sec2} we've described related works in action recognition, data augmentation and video generative model. Section~\ref{methods} introduces our video generation methods as well as skeleton trajectory generation methods with samples and use cases. In \textsection \ref{setup}, we've discussed the data sets and action recognition methods used to evaluate our work. In \textsection \ref{experiments} we've presented the extensive experimental data backing our claims. Our paper is concluded in \textsection \ref{conclude}.


\begin{figure*}[!ht]
\begin{center}
\includegraphics[width=1\linewidth]{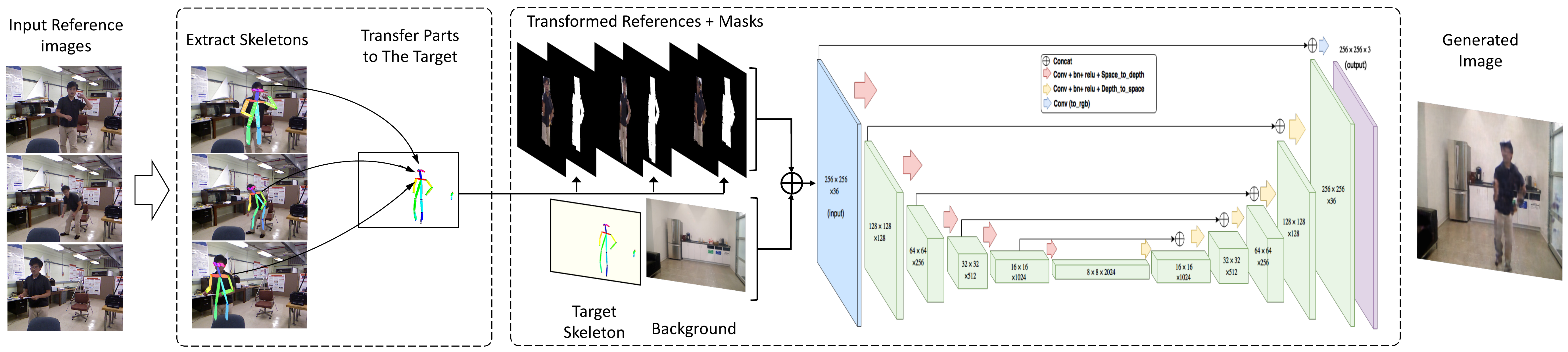}
\end{center}
   \caption{Structure of the network. On the left side "generator network" takes as input background, target skeleton, and the transformed reference images to the target skeleton along with their masks. On the right side "discriminator" takes as input generated image or ground truth and outputs "fake" or "real". }
\label{fig:gen_network}
\vspace{-5mm}
\end{figure*}

\section{Related Works} \label{sec2}

\subsection{Action Recognition}

Human action recognition has drawn attention for some time. Before deep learning era of computer vision, many researchers tried to inflate successful 2D features or descriptors  in order to solve this problem such as 3d SIFT~\cite{scovanner2007}, 3d bag of features~\cite{li2010action} or dense trajectories~\cite{wang2011action}. Please refer to \cite{poppe2010survey} for a comprehensive survey of these types of algorithms.

Deep learning networks significantly outperformed transitional approaches and are therefore the focus of this paper.
Unlike image representation network architecture, the video representation networks haven't had satisfactory advances. There have been different approaches to this problem. Some used the
convolution and layers in 2D (image-based)~\cite{donahue2015long,yue2015beyond} while some used
3D (video-based) kernels~\cite{ji20133d, tran2015learning, carreira2017quo}. Input to the networks could be just RGB video ~\cite{tran2015learning} while optical flow could be used as an additional input~\cite{feichtenhofer2016convolutional, carreira2017quo}. Information could propagate across frames either through LSTMs~\cite{donahue2015long,yue2015beyond} or feature aggregation~\cite{karpathy2014large}. 

\textbf{Data Augmentation}
Using synthetic data or data warping for training classifiers has been proven effective~\cite{krizhevsky2012imagenet, zhang2015learning, simard2003best}. Sato~\etal~\cite{sato2015apac} proposes a method for training a neural network classifier using augmented data. Wong~\etal~\cite{wong2016understanding} thoroughly investigated the benefits of data augmentation for classification tasks. In action recognition tasks, data is usually very limited, since collecting and annotating videos is difficult. Although one can use our algorithm for data augmentation by generating videos varying in background, human appearance, and type of actions, this is not the purpose of our work. Unlike data augmentation that is limited to manipulating data, our method is capable of generating new data with new content and visual features. 

\subsection{Video Generative Models}
Video generation has posed as a challenge for a number of years. The  early work in the field focused on generating texture ~\cite{doretto2003dynamic, szummer1996temporal, wei2000fast}. In recent years with the success of generative models in image generation such as GANs~\cite{goodfellow2014generative}, VAEs~\cite{kingma2013auto, rezende2014stochastic}, Plug\&Play Generative Networks~\cite{nguyen2016plug}, Moment Matching Networks~\cite{li2015generative}, and. PixelCNNs~\cite{van2016conditional}, a new window of opportunity has opened towards generating videos using generative models. In this paper, we use GANs to generate human skeleton trajectories and realistic video sequences. GAN consists of a discriminator and a generator, trained in a 2-player zero-sum game. Although GANs have shown promising results on image generation~\cite{denton2015deep, radford2015unsupervised, zhang2016stackgan, liu2016coupled, liu2017unsupervised}, they have proven to be difficult to train. To address this issue, Arjovsky~\etal~\cite{arjovsky2017wasserstein} proposed Wassertein GAN to combat mode collapse with more stability. Salimans~\etal~\cite{salimans2016improved} introduced several tricks for training GANs. Karras~\etal~\cite{karras2017progressive} proposed a novel method for training GANs through progressively adding new layers.
Ronneberger~\etal~\cite{ronneberger2015u} proposed U-Net, a convolutional network for segmentation.

GANs have previously been used for video generation. There are two lines of work in video generation. First is video prediction where given the first few frames of a video, the goal is to predict the future frames. Several papers focus on producing pixel values conditioned on the past observed frames~\cite{xue2016visual, srivastava2015unsupervised, oh2015action, mathieu2015deep, kalchbrenner2016video, xue2016probabilistic,  villegas2017decomposing}. Another group of papers aimed at reordering the pixels from the previous frames to generate the new ones~\cite{van2017transformation, finn2016unsupervised}.

In the second line of work, the goal is to generate a sequence of video frames conditioned on label, single frame, etc. Early attempts assumed video clips to be fixed length and embedded in a latent space~\cite{vondrick2016generating, saito2016temporal}. Tulyakov~\etal~\cite{tulyakov2017mocogan} proposed to decompose motion from content and generate videos using a recurrent neural net. Our work is different from \cite{tulyakov2017mocogan} where their model learns motion and content in the same network whereas we separated them completely.Furthermore, \cite{tulyakov2017mocogan} is not capable of generating complex human motions. Also filling gaps in the background initially blocked by the person in the input video is a difficult task for this method. On the other hand, our method handles these challenges by completely separating appearance, background, and motion. Our work is somewhat similar to \cite{walker2017pose}, which does video forecasting using pose estimation, by modeling the movement of human using a VAE and then using a GAN to predict the pixel value of the future frames.

\begin{figure}[!b]
\vspace{2mm}
\begin{center}
\includegraphics[width=0.85\linewidth,height=3.3cm]{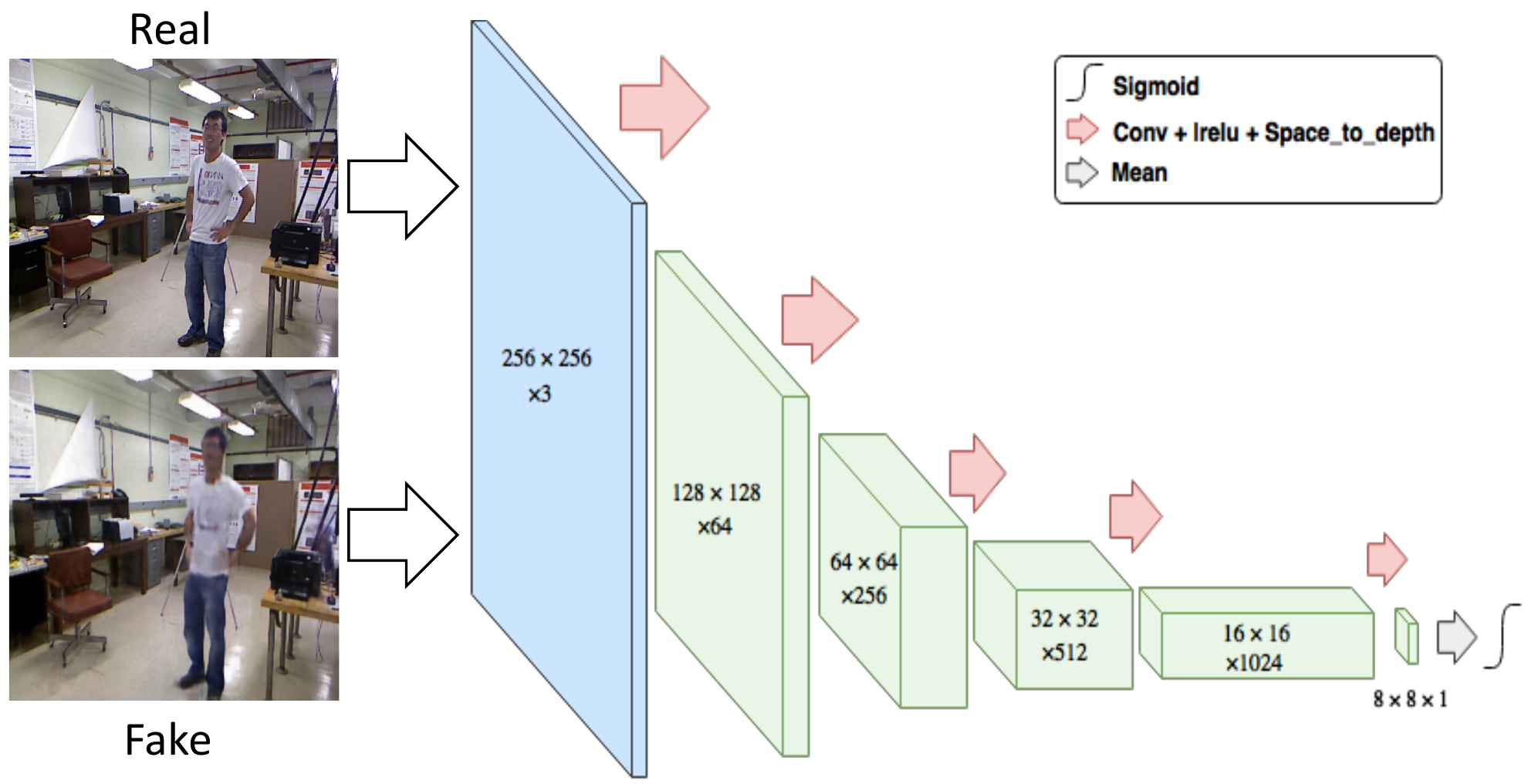}
\end{center}
\caption{Architecture of the discriminator, $D$.}
\label{fig:dis_net}
\end{figure}

Our work lies in the "video generation" category where we focus on employing video generation techniques to generate human action videos. In our proposed method we completely separate background, skeleton motion, and appearance, allowing us to model frame generation and skeleton trajectory independently. So, one would require labeled data and the other can benefit from unlimited unlabeled human action videos available on internet, respectively.

\section{Method} \label{methods}

We define problem as follows; given an action label $l$ a small set of reference images $I = \{I_1, ... I_k\}$ each containing a human subject from which a sequence of video frames is generated featuring a human with the same appearance as the human in the reference image set $I$ performing an action $l$. Modeling the (human/camera) motion and generating photo-realistic video frames may be challenging but knowing the location/motion of human skeletons in each frame would simplify it. Hence, we subdivided the problem into two simpler tasks (inspired by \cite{tulyakov2017mocogan, villegas2017decomposing}). 

\begin{itemize}
\item 
    The first task comprised of the reference images $I$, background image $B$, and a sequence of target skeletons $S=[S_1, S_2, ..., S_n]$ employed to render photo-realistic video frames of the person in $I$ moving according to $S$ on background.
    \item 
    The second task produced the target skeleton sequences for the first part. In another words, given action label $l$, a sequence of skeletons of a random person performing action $l$ was generated.
\end{itemize}
    
By combining the two tasks, we created a novel algorithm that can generate arbitrary number of human action videos with varying backgrounds, human appearances, actions, and ways each action is performed.

\subsection{Video Generation from Skeleton and Reference Appearance} \label{method_generation}
In this section, we explain our algorithm used to generate a video sequence of a person based on given appearance ($I$) and a series of target skeletons ($S$) in an arbitrary background($B$). In our proposed model, we use GAN conditioned on the appearance, the target skeleton, and the background. Our proposed generator network works in a frame-by-frame fashion, where each frame is generated independently from others. We have tried using LSTMs and RNNs to take into account smoothness of the videos. However, our experiments show frames that are generated separately are sharper as RNNs/LSTMS may introduce blurriness to the generated frames.

\textbf{Generator Input}. Our generator network needs a reference image of the person in order to generate images of the same person with arbitrary poses/backgrounds. However, one reference image may not have all the appearance information due to occlusions in some poses (e.g. face is not visible when the person is not facing the camera). To overcome this issue to some extent, we provided multiple reference images of the person to the network. In both training and testing, these images were selected completely at random, so that network would be responsible for choosing the right pieces of appearance features from the set of input images. These images could be selected with a better heuristic to produce better results though this is not in the scope of this work.

\begin{figure}[!ht]

\begin{subfigure}[b]{1\linewidth}
\centering
   \includegraphics[width=0.88\linewidth, height=6cm]{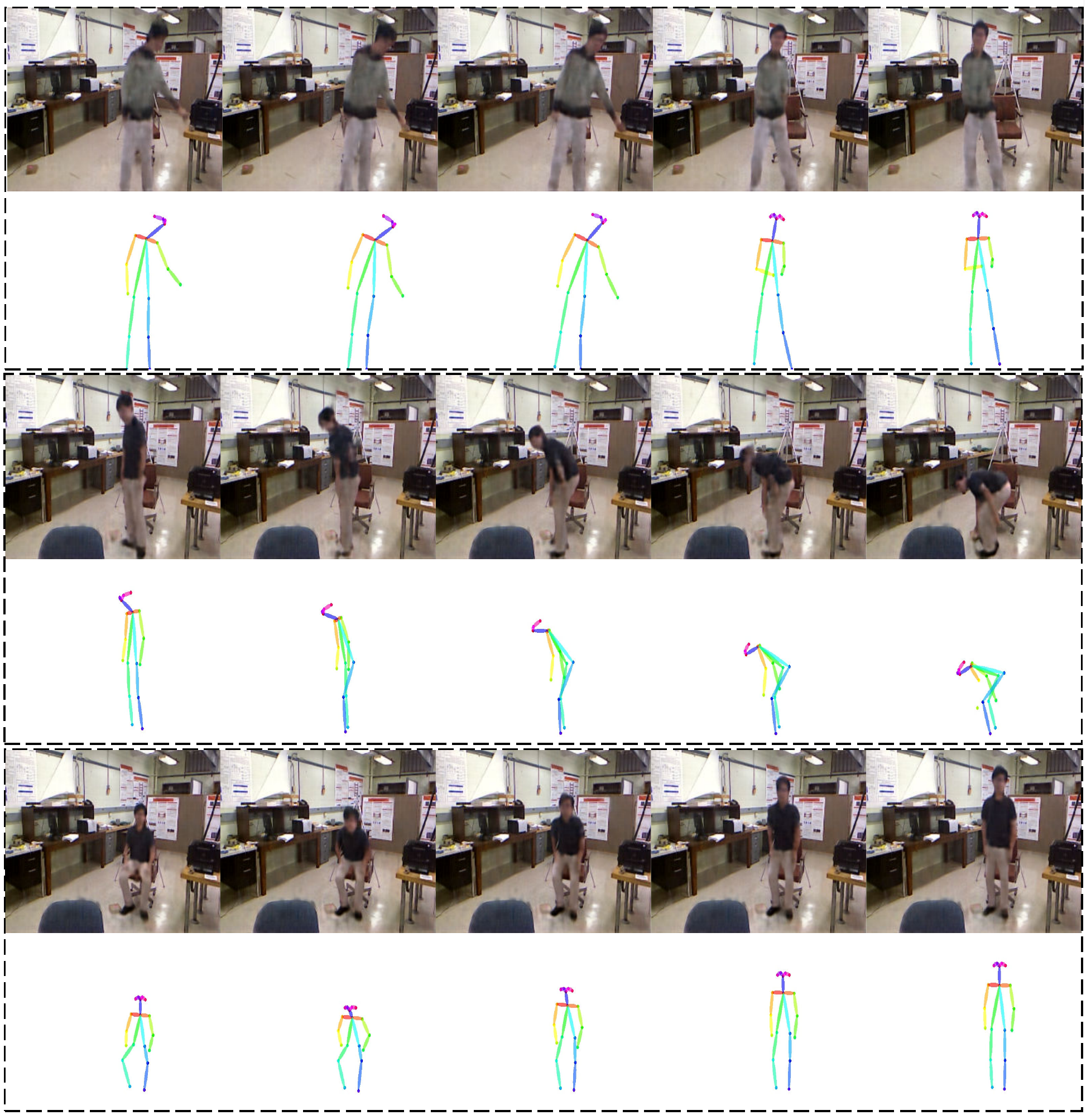}
   \caption{UT dataset. Subjects from the same dataset.}
   \label{fig:ut_results} 
\end{subfigure}

\begin{subfigure}[b]{1\linewidth}
\centering
   \includegraphics[width=0.88\linewidth, height=6cm]{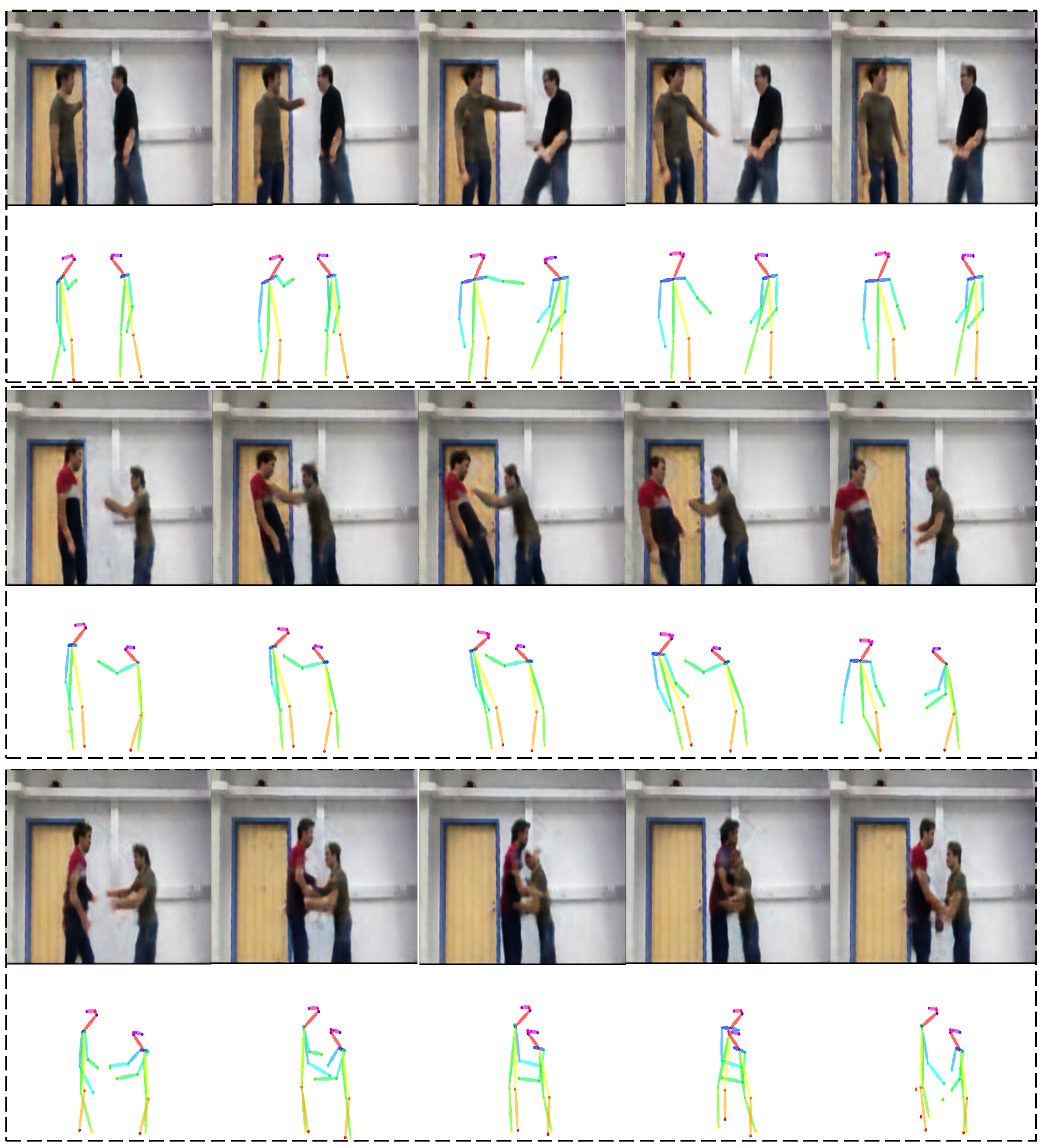}
   \caption{SBU dataset. None of the subjects exist in this dataset.}
   \label{fig:sbu_results}
\end{subfigure}
\caption{Generated images on two different datasets.}
\end{figure}

The reference images were pre-processed before incorporation into the network. First we extracted the human skeleton from each reference image $I_i$ (using \cite{cao2017realtime}), then used an offline transform to map the RGB pixel values of each skeleton part from the image to the target skeleton. Also, a binary mask of where the transformed skeleton is located was created. All these images, $I^t=\{I^t_1, ..., I^t_k\}$, along with the background, $B$, and the target skeleton, $S_i$ were stacked.

\textbf{Conditional GAN.} Inspired by pix2pix~\cite{isola2016image}, we used a U-net style conditional GAN. The generator $G(C)$, is conditioned on the set of transformed images and corresponding masks, along with the background and target skeleton. The generator, $G$, maps $C=\{I^t_1, ..., I^t_k, B, S_i\}$ to the target frame $Y$, such that it fools the discriminator, $D(C, Y)$. The discriminator, $D(C, Y)$, on the other hand is trained to discriminate between real images and the fake images generated by $G$. The architecture of the discriminator is illustrated in Fig.~\ref{fig:dis_net}. The pipeline and architecture of the generator $G$ is illustrated in Fig.~\ref{fig:gen_network}. Fig.~\ref{fig:ut_results} illustrates some of the results.

The objective function of GAN is expressed as:
\setlength{\abovedisplayskip}{1.5pt}
\setlength{\belowdisplayskip}{2pt}
\setlength{\abovedisplayshortskip}{1.5pt}
\setlength{\belowdisplayshortskip}{2pt}

\begin{align*}
    \mathcal{L}_{GAN}\left(G,D\right) &= \mathbb{E}_{c, y \sim P_{data}\left(c,y\right)}[\log D\left(c, y\right)] \\
    &+ \mathbb{E}_{c \sim P_{data}(c), z \sim P_z(z)}[1-\log D\bigl( c, G\left(c, z\right)\bigr) ] 
\end{align*}
Following~\cite{isola2016image} we added an $L1$ loss to the objective function, which resulted in sharper generated frames.
\begin{align*}
    \mathcal{L}_{L1}(G) = \mathbb{E}_{c, y \sim P_{data}(c,y), z \sim P_z(z)}[ || y -  G(c, z) ||] 
\end{align*}
In initial experiments, we noticed that using only $L1$ loss and GAN loss is not enough as the output background would be sharp but the region that the target person is supposed to be was blurry. Subsequently, we introduced a "Regional L1 loss" with a larger weight as following,
\begin{align*}
    \mathcal{L}_{R}(G) = &\mathbb{E}_{c, y \sim P_{data}(c,y), z \sim P_z(z)}\\
    &[ || \text{masked}(y) -  \text{masked}(G(c, z)) ||] 
\end{align*}
where "masked" masks out the region where the person was located. This mask was generated based on the target skeleton, $S_i$, using morphological functions (erode, etc.).

Our final objective is as follows:
\begin{align*}
    \mathcal{L}(G, D) = \mathcal{L}_{GAN}(G, D) + \lambda \mathcal{L}_{L1}(G) + \beta \mathcal{L}_{R}(G)
\end{align*}
where $\lambda$ and $\beta$ are weights of $L1$ and $R$ regional losses (in our experiments $\beta > \lambda$). 
and the goal is to solve the following optimization problem.
\begin{equation}
    G^* = \arg\,\min_G\,{\max_D}\mathcal{L}(G, D)
\end{equation}



\textbf{Multi-person Video Generation}
In a nutshell, our algorithm merges transformed images of a person on an arbitrary pose with an arbitrary background in a natural photo-realistic way. We managed to go beyond simple one person human action videos and extended our method to \textit{multi-person interaction videos} as well.
For this purpose, we trained our model on a two person interaction data set~\cite{yun2012two}. 
The only difference with single frame generation process is that in the pre-processing phase, for each person in the input reference image, we needed to know the corresponding skeleton in the target frame, we then transformed each person's body parts to his/her own body parts in the target skeleton. There are some challenges in this task such as occlusions in certain interactions (e.g. passing by, hugging, etc.). The data set that we used contains these occlusions to some extent. Our method is able to handle  relatively well some simple occlusions that occur in such interactions. We acknowledge that there is room for improvement in this area, but that would not fit in the scope of this work. Fig.~\ref{fig:sbu_results} illustrates some of the generated videos.

\subsection{Skeleton Trajectory Generation} \label{method_trajectory}
    In the previous section, we explained how we designed a method that enables us to generate videos of an arbitrary person in any background based on any given sequence of skeletons. Although number of backgrounds and persons are unlimited, the number of labeled skeleton sequences are limited to the ones in the existing data sets. We propose a novel solution to this problem; using a generative model to learn the distribution of skeleton sequences conditioned on the action labels. This allows us to generate as many skeleton sequences as needed for the actions in the data set. Fig.~\ref{fig:generated_skeletons} shows a few sample generated skeleton sequences.
    
    We used small data sets for training our model. However, due to the nature of the problem and the limited amount of data, generating long sequences of natural looking skeletons proved challenging. Thus we aimed at generating relatively short fixed-length sequences. Having said that, training GAN in such way is still prone to problems such as mode collapse, divergence, etc. In designing the generator and discriminator networks, we have taken into account these problems (e.g. introduced batch diversity in the discriminator, created multiple discriminators, etc.). 
    
    \textbf{Skeleton Trajectory Representation}. 
    Each skeleton consists of 18 joints. We represented each skeleton with a $1\times36$ vector (a flattened version of $18\times2$ matrix of joints coordinates). We normalized the coordinates by dividing them by "height" and "width" of the original image.  
    
    \textbf{Generator Network}. We used a conditional GAN model to generate sequences of skeletal positions corresponding to different actions. Our generator has a "U" shape architecture where input consists of action label and noise, and output is a $8\times1\times36$ tensor representing a human skeleton trajectory with $8$ time-steps.
    
\begin{figure}[!t]
    
    \begin{subfigure}[b]{1\linewidth}
        \centering
       \includegraphics[width=0.75\linewidth,height=2.6cm]{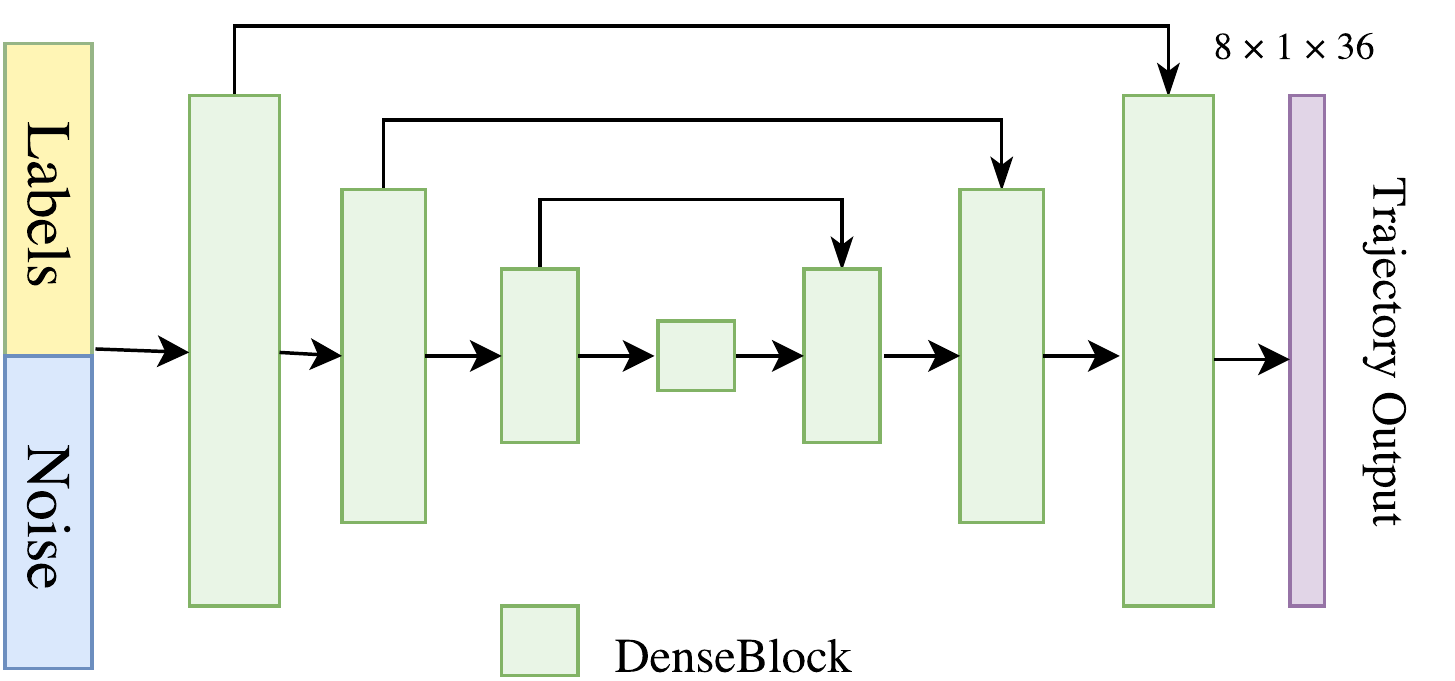}
       \caption{Generator Network.}
       \label{fig:traj_gen}
    \end{subfigure}
    
    \begin{subfigure}[b]{1\linewidth}
    \centering
       \includegraphics[width=0.8\linewidth]{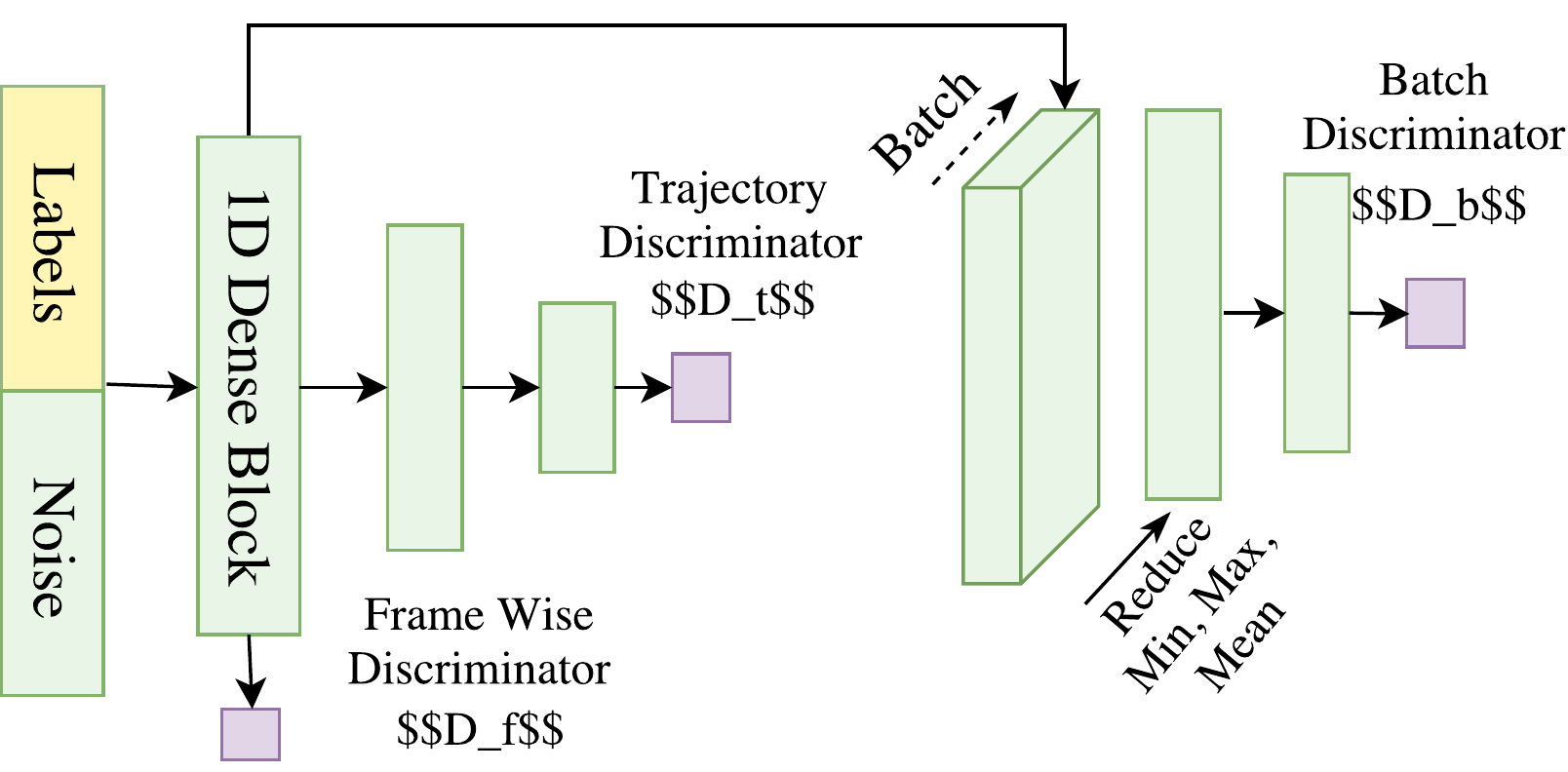}
       \caption{Trajectory Discriminator Network. The discriminator is the sum of three discriminators illustrated in this figure: $D=D_f + D_t + D_f$.}
       \label{fig:traj_dis} 
    \end{subfigure}

    \caption{Trajectory GAN network architecture.}
    \vspace{4mm}
    \end{figure}

    Based on our results, providing a vector of random noise for each time step helps the generator to learn and generalize better. So the input noise, $z$, is a tensor with size $8\times1\times128$; drawn from a uniform distribution. The one-hot encoding of action label, $l$, is replicated and concatenated to the 3rd dimension of the $z$. The rest is a "U" shaped network with skip connections that maps the input ($z,l$) to a skeleton sequence $S$. Fig.~\ref{fig:traj_gen} illustrates the network architecture. We also used Dense-net~\cite{huang2017densely} blocks in our network.
    
    \textbf{Discriminator Network}. Architecture of discriminator is three-fold. The base for discriminator is 1D convolutional neural net along the time dimension. In order to allow discriminator to distinguish "human"-looking skeletons, we used sigmoid layer on top of fully-convolutional net. To discriminate "trajectory", we used set of convolutions along the time with stride 2, shrinking output to one $1\times1\times C$ containing features of the whole sequence. To prevent mode collapse, first we grouped fully convolutional net outputs across batch dimension.We then used min, max and mean operations across batch, and provided these statistical information to the discriminator. This method seems to provide enough information about distribution of values across batch and allows to change batch size during training. For detailed discriminator architecture see Fig.~\ref{fig:traj_dis}.
    
    Our objective function is:

    \begin{figure}[ht!]
\begin{center}
\includegraphics[width=0.80\linewidth,height=5cm]{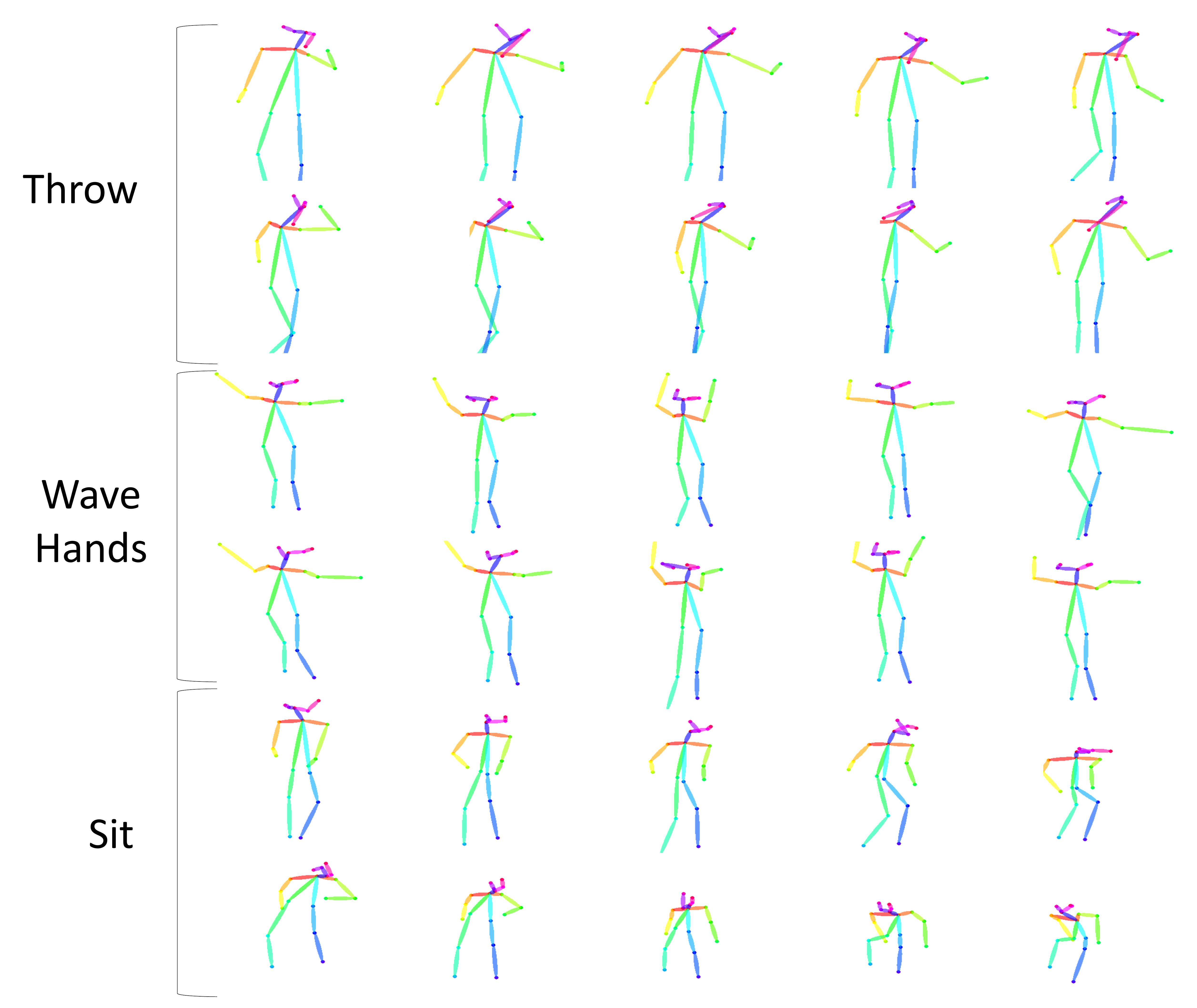}
\end{center}
   \caption{Samples of generated skeleton sequences, conditioned on action label (e.g. throwing, hand waving, sitting). 
   \vspace{4mm}
   }
\label{fig:generated_skeletons}

\end{figure}

    \begin{align*}
    \mathcal{L}_{T}\left(G,D\right) &= \mathbb{E}_{l, s \sim P_{data}\left(l,s\right)}[\log D\left(l, s\right)] \\
    &+ \mathbb{E}_{l \sim P_{data}(l), z \sim P_z(z)}[1-\log D\bigl( l, G\left(l, z\right)\bigr) ] 
    \end{align*}
    
    where $l$ and $s$ are action label and skeleton trajectories, respectively. We aim to solve the following:
    
    \begin{align*}
    G^* = \arg\,\min_G\,{\max_D}\mathcal{L}_T(G, D)
    \end{align*} 
     
    In this work, we have shown that generative models can be adopted to learn human skeleton trajectories. We trained a Conditional GAN on a very small data set (200 sequences) and managed to generate natural looking skeleton trajectories conditioned on action labels.  This can be used to generate a variety of human action sequences that don't exist in the data set.
    However, our work is limited to a fixed number of frames. Thus for future work, we'll work to improve our method so that it'll accommodate longer sequences varying in length.
    We also explained that in addition to the generated skeletons, we can also use real skeleton sequences from other sources (other data sets, current data set but different subjects) to largly expand existing data sets.

\section{Datasets and Action Recognition Methods} \label{setup}

\subsection{Data Sets} \label{dataset}
    In this paper, we've claimed to expand small amount of action videos by addition of new generated videos. We targeted smaller action recognition data sets and expanded them to meet the large data load requirements of recent action recognition algorithms such as UCF 101~\cite{soomro2012ucf101}, the kinetics~\cite{kay2017kinetics} or NTU RGB+D~\cite{shahroudy2016ntu}. This eliminates the need for time and cost inefficient data acquisition processes.  
    
\textbf{UT Kinects~\cite{xia2012view}:} One of the data sets wildly used in our experiments is UT Kinects which includes 10 action labels: Walk, Sit-down, Stand-up, Trow, Push, Pull, Wave-hand, Carry and Clap-hand. There are 10 subjects that perform each of these action twice in front of a rig of RGB camera and Kinect. Therefore in total they are 200 action clips of RGB and depth though depth is ignored. All videos are taken in office environment with similar lighting condition and the position of the camera is fixed.

For the training setup, 2 random subjects were left out (20\%, used for testing) and the experiments were carried out using 80\% of the subjects. The reported results are the average of  six individual runs. The 6 train/test runs are constant throughout our experiment.           

\textbf{SUB Interact~\cite{yun2012two}:} Since our methods work with multiple human subjects in a scene, we picked SUB Interact. It is a kinect captured human activity recognition data set depicting two person interaction. It contains 294 sequences of 8 classes (Kicking, Punching, Hugging, Shaking-hand, Approaching, departing and Exchanging objects) with subject independent 5-fold cross validation. The original data includes RGB, depth and skeleton but we only use RGB for our purpose. We used a 5-fold cross validation throughout our experiments and reported the average accuracy.    

\textbf{KTH~\cite{kth2004}:} KTH action recognition data set was commonly used at the early stage of action recognition. It includes 600 low resolution clips of 6 actions: Walk, Wave-hand, Clap-hand, Jogging, running and boxing which are divided in train, test and validation. The first three action labels are shared with UT data set while the last three are new. We used this data set to add new action to UT data set and for cross data set evaluation. 

\subsection{Action Recognition Methods}

We used the following deep learning networks which have previously shown decent performance on recent action recognition data sets.

\textbf{Convolutional 3D (C3D)~\cite{tran2015learning}:} is a simple and efficient 3-dimensional ConvNet for spatiotemporal feature which shows decent performance on video processing benchmarks such as action recognition in conjunction with large amount of training data. We used their proposed network with 8 convolutional layers, 5 pooling layers and 2 fully connected layers with 16-frames of $112\times112$ RGB input. They released a network pre-trained on UCF~Sport~\cite{soomro2012ucf101} which we used for our experiments aimed at training from scratch, denoted as C3D(p) vs. C3D(s). Unfortunately we can not couldn't converge the C3D when we trained from scratch on UT data set but it converged successfully on SUB. 

\textbf{Inflated 3D ConvNets (I3D)~\cite{carreira2017quo} :} is a more complex model which has recently been proposed as the state-of-the-art for action recognition task. It builds upon Inception-v1~\cite{ioffe2015batch}, but inﬂates their filters and pooling kernels into 3D. It is a two-steam network which uses both RGB and optical flow input with $224 \times 224$ inputs. We only used RGB for simplicity. They released a network pre-trained on  ImgeNet~\cite{deng2009imagenet} followed by the Kinetics~\cite{kay2017kinetics}. We used this for our experiments aimed at training from scratch, denoted as I3D(p) vs. I3D(s).

We use data augmentation by translation and clipping as mentioned in~\cite{carreira2017quo} for all experiments. For training, we only used the original clips as test, making sure there was no generated clips with skeletons or subjects (subject pair) from test data in each run.

\section{Experiments} \label{experiments}
So far, we have introduced our video generation method which enable us to generate new action clips for the action recognition training process. In this section, we show different scenarios for generating new data and running experiments for each to see if adding the generated data to a training process can improve the accuracy of the action recognizer. We applied our proposed video generation models to all the experiments using skeletons. The skeletons were trained using data from UT and SBU data sets as well as 41 un-annotated clips (between 10 to 30 seconds) that we captured from our colleagues. For future works, we will train our model again using a large amount of data from web. But the time being,  we are satisfied with the current model as higher resolution for action recognition is currently unnecessary. Our technique for generating new action video clips has the capacity of running experiments with numerous varying settings. Here, we show five experiments which may be quantitatively evaluated. 

\subsection{Generated Trajectory}

The first experiments is a combination of our proposed video generation technique and skeleton trajectory generation. We generated around 200 random skeleton trajectories from action labels in UT data set using the method mentioned in \textsection \ref{method_trajectory}. Each of these skeleton trajectories generated a video by proposed video generation applied to a person in UT data set, meaning our new data set is doubled with half of it being the generated data. We then trained our model by I3D and C3D using training setting mentioned in \textsection \ref{dataset}. Table~\ref{tab:gen} shows about 3\% improvement for I3D with and without training data as well as significant improvement (by 15\%) for C3D network which is less complex.    
\begin{table}[!htbp]
\begin{center}
\begin{tabular}{l|c c}
 Method & Org. & Org. + Gen. \\
\hline \hline
I3D(s) & 64.58\% & 67.50\% \\
I3D(p) & 86.25\% & 89.17\%  \\
C3D(p) & 55.83\% & 70.83\%   \\
\hline
\end{tabular}
\end{center}
\caption{ Action recognition on UT data set using original data compared to generated from scratch data with proposed method in \textsection \ref{method_generation} and  \textsection \ref{method_trajectory}  }
\vspace{-8mm}
\label{tab:gen}
\end{table}

\subsection{New Subjects}
One common way to extend a video data set is to invite new people to do a series of actions in front of a camera. Diversity~\cite{bagheri2015keep} in body shape, cloths and behaviour will clearly help with the generalization of the ML methods. In this experiment, we aimed to virtually add new subject to the data set. Thus, we collected a small unannotated clips from 10 distinct persons and fed them as new subjects into our proposed video generation method. For UT, each subject was replaced by a new one for all of his/her action which is similar to adding 10 new subjects to UT. The same was done with SUB to double the data set, the only difference being the replacement each pair with a new subject pair. Figure~\ref{fig:sbu_results} shows a few new subjects with their generated action videos from SBU data set. The results have been presented in Table~\ref{tab:newsubj}.

\begin{table}[!htbp]
\begin{center}
\begin{tabular}{l|c c|c c}
 & \multicolumn{2}{c}{UT} & \multicolumn{2}{c}{SBU} \\
 & Org. & Exp. & Org & Exp. \\
\hline\hline
I3D(s) &     64.58\% & 67.08\% & 86.48\% & 91.23\%  \\
I3D(p) & 86.25\% & 89.17\% &          97.30\% & 98.65\% \\
C3D(s) & - & - &         83.52\% & 87.00\%   \\
C3D(p) &  55.83\% & 70.43\%  &          92.02\% & 96.25\%  \\

\hline
\end{tabular}
\end{center}
\caption{ Performance comparison of multiple algorithms, trained on original data and additional subjects.}
\label{tab:newsubj}
\vspace{4mm}
\end{table}

\begin{figure}[!hb]
\begin{center}
\includegraphics[width=0.85\linewidth,height=6cm]{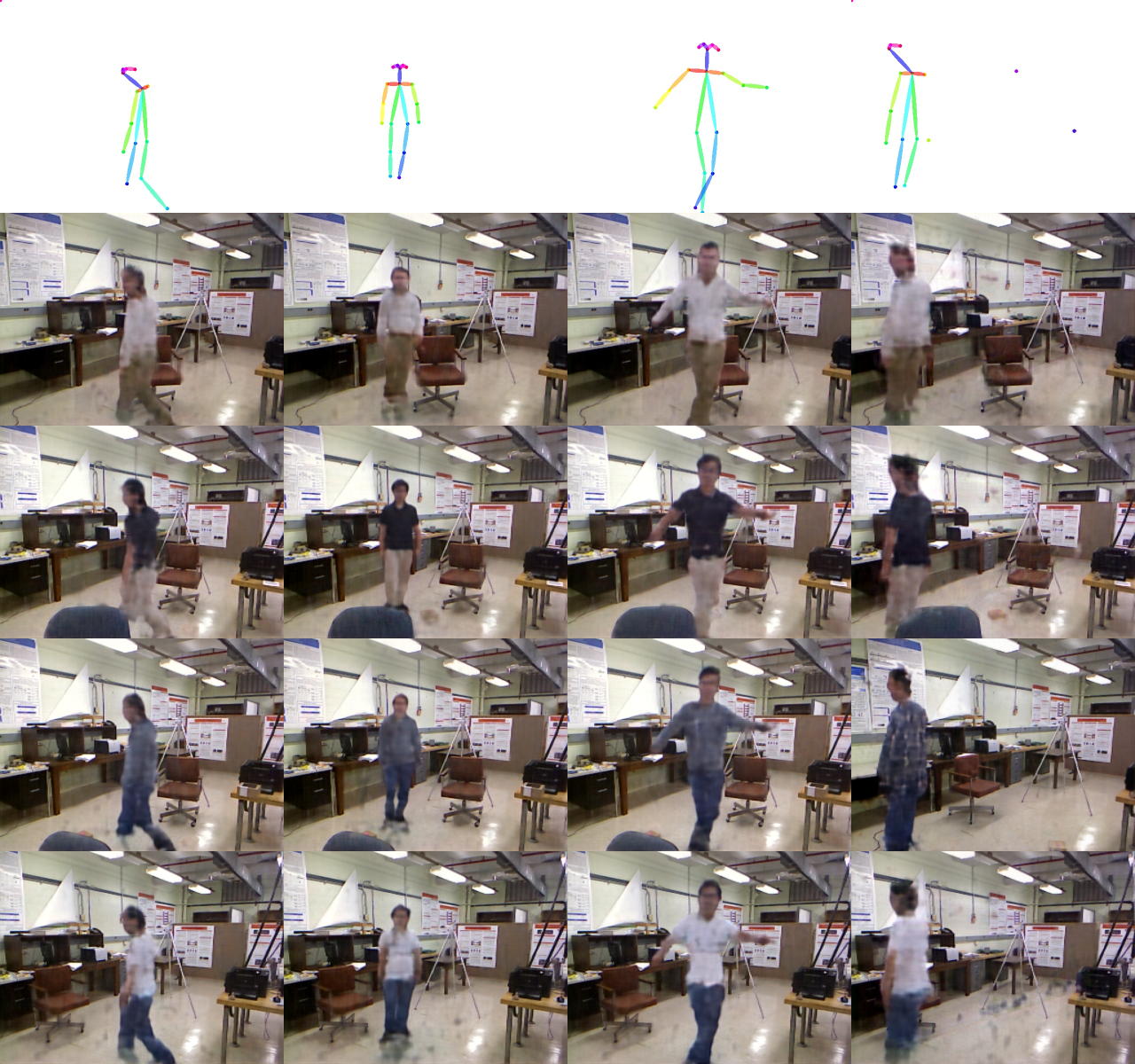}
\end{center}
   \caption{The screen shot of a video generated by UTK expansion. The first row shows skeleton clips extracted from an arbitrary action. Second to fourth rows show the generated video for subjects from different clip carrying out that specific action.}
\label{fig.expanded_ut}
\end{figure}

\subsection{New Actions}

In real computer vision problems, one might decide to add a new label class after the data collection process has been done. Adding a new label action to a valid data set could cost the same as gathering a data set from scratch as all the subjects are needed for re-acting that single action. In this experiment, we tried to introduce a new action labeled to UT data set. As mentioned in \textsection \ref{dataset} , UT consists 10 action labels. We used training data from a third data set called KTH~\cite{kth2004} in order to generate 3 new actions, running, jogging and boxing, in addition to that of the UT. For each subject in UT data set and each of these 3 new action, we randomly picked 5 action clips from KTH training data clips and extracted the skeleton by OpenPose~\cite{cao2017realtime} where in addition to input background image, we generated 150 new action clips from our data set. We then trained a new model using I3D by pre-trained network where in each run we used training data from original set and all the data generated for the new set of actions. Since the KTH data is grey scaled images, we randomly grey scaled both the original and the generated training clips in the training phase. For each run, we found per class accuracy for UT test set (refer to \textsection \ref{dataset} for explaining UT train/test) as well as KTH test sets. Table~\ref{tab:newact} shows average of the per class accuracy for both test sets. We may consider KTH test results as a measure  of cross data set accuracy for walk, wave-hand and clap-hand. Our trained network on new action labels \textit{boxing}, \textit{running} and \textit{jogging} achieved 72.14\%, 44.44\% and 63.20\%, respectively. This indicates that the new actions in the data set performed as good as the data captured by camera.  

\begin{table}[!htbp]
\begin{center}
\begin{tabular}{l|c||l|c}
 Action & UTK Test & Label & KTH Test \\
\hline\hline
Walk &     91.67\% &       Walk & 67.18\%   \\
Wave-hand &     100.0\% &  Wave-hand & 58.59\%   \\
Clap-hand &     91.67\% &  Clap-hand & 28.90\%   \\
Push &     33.33\% &       \textbf{Boxing} & \textbf{72.14\% }  \\
Pull &     58.33\% &       \textbf{Running} & \textbf{44.44\%}   \\
Pick-up &     100.0\% &    \textbf{Jogging} & \textbf{63.20\%}   \\
Sit-down & 87.50\% & &   \\
Stand-up & 95.83\% &     & \\
Threw &  54.17\% & &   \\
Carry &  79.17\% & &   \\

\end{tabular}
\end{center}
\caption{ Per class average accuracy for model trained by i3d using original training data from UT plus new action clip generated by our method using skeleton extracted from KTH training set. }
\label{tab:newact}
\vspace{-4mm}
\end{table}

\subsection{Data set Expansion}

So far, we've shown that using our proposed method we can generate video clips with any number of arbitrary action videos and subjects. In an action data set with $N$ subjects carrying out $M$ distinct actions, there will be $M\times N$ video actions. when applied to our proposed method of action video generation, the $N$ subjects and the $M\times N$ video actions will result in generation of $M\times N^2$ video actions comprising of $M\times N$ original videos while the rest is generated videos. This approach enabled us to expand UT Kinect data set from 200 clips to 4000 clips and SUB Interact from 283 clips to 5943 using only the original data set. We trained I3D and C3D using our expanded data set as described in \textsection \ref{dataset}. Table~\ref{tab:expanded} shows the result of this experiment.  

\begin{table}[!htbp]
\begin{center}
\begin{tabular}{l|c c|c c}
 & \multicolumn{2}{c}{UTK} & \multicolumn{2}{c}{SBU} \\
 & Org. & Exp. & Org & Exp. \\
\hline\hline
i3d(s) &     64.58\% & 69.58\% & 86.48\% & 93.54\%  \\
i3d(p) & 86.25\% & 90.42\% &          97.30\% & 99.13\% \\
c3d(s) & - & - &         83.52\% & 86.03\%   \\
c3d(p) &  55.83\% & 71.25\%  &          92.02\% & 97.41\%  \\

\hline
\end{tabular}
\end{center}
\caption{ The comparison of data set expansion by original data for UTK and SUB data set.}
\label{tab:expanded}
\end{table}

Figures~\ref{fig.expanded_ut} shows an screen shot of the clips from UTK and SUB data sets. The first row shows skeleton clips extracted from an arbitrary action while rows 2-4 show the generated video for subjects from different clip performing that specific action.

\subsection{Real World}

\begin{figure}[ht!]
\begin{center}
\includegraphics[width=0.9\linewidth]{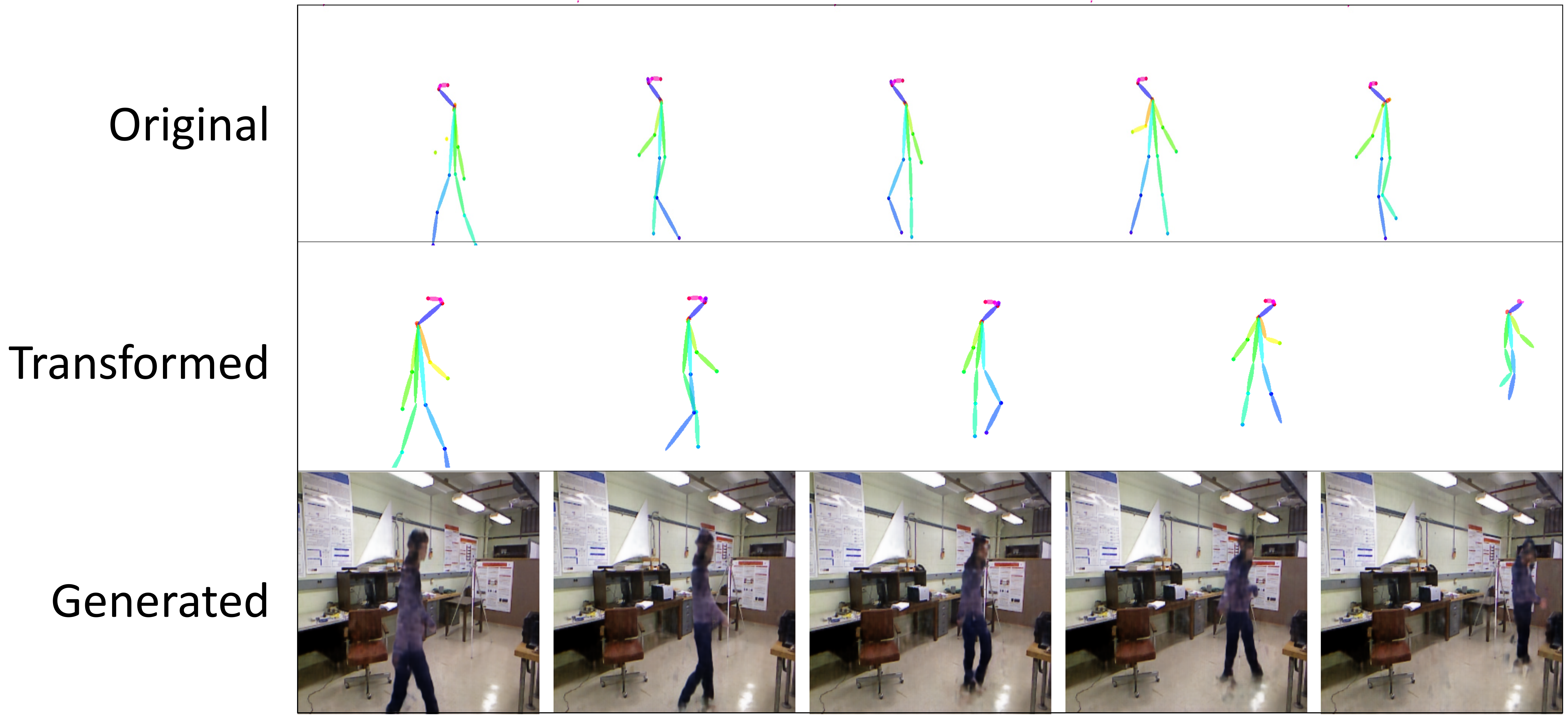}
\end{center}
   \caption{Perspective transform example.}
\label{fig.perspective}
\end{figure}

In this section, we carried out 4 different experiments on 2 data sets for bench-marking. Although in all experiments, the generated data improved the network performance, we believe none of the experiments show the actual strength and convenience of our proposed methods in real world scenarios. In both data sets, as well as other commonly used small data sets, the environmental setup for data acquisition such as distance from camera view~\cite{junejo2011view} and light condition were kept as uniformly as possible for both test and train video clips. This would be unattainable in real life data acquisitions. A way of overcoming this obstacle would be to collect diverse sets of data for strong neural network models. We've previously shown that by partitioning the video to action, subject and context allows us to easily manipulate the background or change the camera view. In this experiment, We applied perspective transform on skeleton while using diverse backgrounds. Although the model trained with these data did not outperform our previous experiments, a live demo showed it to be better for unseen cases, qualitatively. Figure~\ref{fig.perspective} illustrates an input skeleton and its perspective transform as well as the generated clip.

%

\section{Conclusion and Future Works} \label{conclude}

 In this paper, we've introduced a novel way to partition an action video clip into action, subject and context. We showed that we can manipulate each part separately, reassemble them with our proposed video generation model into new clips and use as an input for action recognition models which require large data. We can change an action by extracting it from an arbitrary video clip, generate it through our proposed skeleton trajectory model or by applying perspective transform on existing skeleton. Additionally, we can change the subject and the context using arbitrary video clips.
 
 For the future work, we will replace our 2d skeleton with 3d skeleton to achieve a 3d transformation and handle occlusions. Additionally, while our video generation technique demonstrated acceptable results for $255\times255$ images, we believe it can be extended even further to achieve higher resolution by feeding more unannotated data.   

{\small
\bibliographystyle{ieee}
\bibliography{egbib}

\begin{thebibliography}{10}\itemsep=-1pt

\bibitem{arjovsky2017wasserstein}
M.~Arjovsky, S.~Chintala, and L.~Bottou.
\newblock Wasserstein gan.
\newblock {\em arXiv preprint arXiv:1701.07875}, 2017.

\bibitem{bagheri2015keep}
M.~Bagheri, Q.~Gao, S.~Escalera, A.~Clapes, K.~Nasrollahi, M.~B. Holte, and
  T.~B. Moeslund.
\newblock Keep it accurate and diverse: Enhancing action recognition
  performance by ensemble learning.
\newblock In {\em CVPRW}, pages 22--29, 2015.

\bibitem{cao2017realtime}
Z.~Cao, T.~Simon, S.-E. Wei, and Y.~Sheikh.
\newblock Realtime multi-person 2d pose estimation using part affinity fields.
\newblock In {\em CVPR}, 2017.

\bibitem{carreira2017quo}
J.~Carreira and A.~Zisserman.
\newblock Quo vadis, action recognition? a new model and the kinetics dataset.
\newblock {\em arXiv preprint arXiv:1705.07750}, 2017.

\bibitem{deng2009imagenet}
J.~Deng, W.~Dong, R.~Socher, L.-J. Li, K.~Li, and L.~Fei-Fei.
\newblock Imagenet: A large-scale hierarchical image database.
\newblock In {\em Computer Vision and Pattern Recognition, 2009. CVPR 2009.
  IEEE Conference on}, pages 248--255. IEEE, 2009.

\bibitem{denton2015deep}
E.~L. Denton, S.~Chintala, R.~Fergus, et~al.
\newblock Deep generative image models using a￼ laplacian pyramid of
  adversarial networks.
\newblock In {\em Advances in neural information processing systems}, pages
  1486--1494, 2015.

\bibitem{donahue2015long}
J.~Donahue, L.~Anne~Hendricks, S.~Guadarrama, M.~Rohrbach, S.~Venugopalan,
  K.~Saenko, and T.~Darrell.
\newblock Long-term recurrent convolutional networks for visual recognition and
  description.
\newblock In {\em CVPR}, pages 2625--2634, 2015.

\bibitem{doretto2003dynamic}
G.~Doretto, A.~Chiuso, Y.~N. Wu, and S.~Soatto.
\newblock Dynamic textures.
\newblock {\em International Journal of Computer Vision}, 51(2):91--109, 2003.

\bibitem{feichtenhofer2016convolutional}
C.~Feichtenhofer, A.~Pinz, and A.~Zisserman.
\newblock Convolutional two-stream network fusion for video action recognition.
\newblock In {\em Proceedings of the IEEE Conference on Computer Vision and
  Pattern Recognition}, pages 1933--1941, 2016.

\bibitem{finn2016unsupervised}
C.~Finn, I.~Goodfellow, and S.~Levine.
\newblock Unsupervised learning for physical interaction through video
  prediction.
\newblock In {\em Advances in Neural Information Processing Systems}, pages
  64--72, 2016.

\bibitem{goodfellow2014generative}
I.~Goodfellow, J.~Pouget-Abadie, M.~Mirza, B.~Xu, D.~Warde-Farley, S.~Ozair,
  A.~Courville, and Y.~Bengio.
\newblock Generative adversarial nets.
\newblock In {\em Advances in neural information processing systems}, pages
  2672--2680, 2014.

\bibitem{huang2017densely}
G.~Huang, Z.~Liu, L.~van~der Maaten, and K.~Q. Weinberger.
\newblock Densely connected convolutional networks.
\newblock In {\em Proceedings of the IEEE Conference on Computer Vision and
  Pattern Recognition}, 2017.

\bibitem{ioffe2015batch}
S.~Ioffe and C.~Szegedy.
\newblock Batch normalization: Accelerating deep network training by reducing
  internal covariate shift.
\newblock In {\em ICML}, pages 448--456, 2015.

\bibitem{isola2016image}
P.~Isola, J.-Y. Zhu, T.~Zhou, and A.~A. Efros.
\newblock Image-to-image translation with conditional adversarial networks.
\newblock {\em arXiv preprint arXiv:1611.07004}, 2016.

\bibitem{ji20133d}
S.~Ji, W.~Xu, M.~Yang, and K.~Yu.
\newblock 3d convolutional neural networks for human action recognition.
\newblock {\em PAMI}, 35(1):221--231, 2013.

\bibitem{junejo2011view}
I.~N. Junejo, E.~Dexter, I.~Laptev, and P.~Perez.
\newblock View-independent action recognition from temporal self-similarities.
\newblock {\em PAMI}, 33(1):172--185, 2011.

\bibitem{kalchbrenner2016video}
N.~Kalchbrenner, A.~v.~d. Oord, K.~Simonyan, I.~Danihelka, O.~Vinyals,
  A.~Graves, and K.~Kavukcuoglu.
\newblock Video pixel networks.
\newblock {\em arXiv preprint arXiv:1610.00527}, 2016.

\bibitem{karpathy2014large}
A.~Karpathy, G.~Toderici, S.~Shetty, T.~Leung, R.~Sukthankar, and L.~Fei-Fei.
\newblock Large-scale video classification with convolutional neural networks.
\newblock In {\em Proceedings of the IEEE conference on Computer Vision and
  Pattern Recognition}, pages 1725--1732, 2014.

\bibitem{karras2017progressive}
T.~Karras, T.~Aila, S.~Laine, and J.~Lehtinen.
\newblock Progressive growing of gans for improved quality, stability, and
  variation.
\newblock {\em arXiv preprint arXiv:1710.10196}, 2017.

\bibitem{kay2017kinetics}
W.~Kay, J.~Carreira, K.~Simonyan, B.~Zhang, C.~Hillier, S.~Vijayanarasimhan,
  F.~Viola, T.~Green, T.~Back, P.~Natsev, et~al.
\newblock The kinetics human action video dataset.
\newblock {\em arXiv preprint arXiv:1705.06950}, 2017.

\bibitem{kemelmacher2016megaface}
I.~Kemelmacher-Shlizerman, S.~M. Seitz, D.~Miller, and E.~Brossard.
\newblock The megaface benchmark: 1 million faces for recognition at scale.
\newblock In {\em CVPR}, pages 4873--4882, 2016.

\bibitem{kingma2013auto}
D.~P. Kingma and M.~Welling.
\newblock Auto-encoding variational bayes.
\newblock {\em arXiv preprint arXiv:1312.6114}, 2013.

\bibitem{krizhevsky2012imagenet}
A.~Krizhevsky, I.~Sutskever, and G.~E. Hinton.
\newblock Imagenet classification with deep convolutional neural networks.
\newblock In {\em NIPS}, pages 1097--1105, 2012.

\bibitem{li2010action}
W.~Li, Z.~Zhang, and Z.~Liu.
\newblock Action recognition based on a bag of 3d points.
\newblock In {\em CVPRW}, pages 9--14. IEEE, 2010.

\bibitem{li2015generative}
Y.~Li, K.~Swersky, and R.~Zemel.
\newblock Generative moment matching networks.
\newblock In {\em Proceedings of the 32nd International Conference on Machine
  Learning (ICML-15)}, pages 1718--1727, 2015.

\bibitem{lin2014microsoft}
T.-Y. Lin, M.~Maire, S.~Belongie, J.~Hays, P.~Perona, D.~Ramanan,
  P.~Doll{\'a}r, and C.~L. Zitnick.
\newblock Microsoft coco: Common objects in context.
\newblock In {\em ECCV}, pages 740--755. Springer, 2014.

\bibitem{liu2017unsupervised}
M.-Y. Liu, T.~Breuel, and J.~Kautz.
\newblock Unsupervised image-to-image translation networks.
\newblock {\em arXiv preprint arXiv:1703.00848}, 2017.

\bibitem{liu2016coupled}
M.-Y. Liu and O.~Tuzel.
\newblock Coupled generative adversarial networks.
\newblock In {\em Advances in neural information processing systems}, pages
  469--477, 2016.

\bibitem{marszalek09}
M.~Marsza{\l}ek, I.~Laptev, and C.~Schmid.
\newblock Actions in context.
\newblock In {\em CVPR}, 2009.

\bibitem{mathieu2015deep}
M.~Mathieu, C.~Couprie, and Y.~LeCun.
\newblock Deep multi-scale video prediction beyond mean square error.
\newblock {\em arXiv preprint arXiv:1511.05440}, 2015.

\bibitem{nguyen2016plug}
A.~Nguyen, J.~Yosinski, Y.~Bengio, A.~Dosovitskiy, and J.~Clune.
\newblock Plug \& play generative networks: Conditional iterative generation of
  images in latent space.
\newblock {\em arXiv preprint arXiv:1612.00005}, 2016.

\bibitem{oh2015action}
J.~Oh, X.~Guo, H.~Lee, R.~L. Lewis, and S.~Singh.
\newblock Action-conditional video prediction using deep networks in atari
  games.
\newblock In {\em Advances in Neural Information Processing Systems}, pages
  2863--2871, 2015.

\bibitem{poppe2010survey}
R.~Poppe.
\newblock A survey on vision-based human action recognition.
\newblock {\em Image and vision computing}, 28(6):976--990, 2010.

\bibitem{radford2015unsupervised}
A.~Radford, L.~Metz, and S.~Chintala.
\newblock Unsupervised representation learning with deep convolutional
  generative adversarial networks.
\newblock {\em arXiv preprint arXiv:1511.06434}, 2015.

\bibitem{rezende2014stochastic}
D.~J. Rezende, S.~Mohamed, and D.~Wierstra.
\newblock Stochastic backpropagation and variational inference in deep latent
  gaussian models.
\newblock In {\em International Conference on Machine Learning}, 2014.

\bibitem{ronneberger2015u}
O.~Ronneberger, P.~Fischer, and T.~Brox.
\newblock U-net: Convolutional networks for biomedical image segmentation.
\newblock In {\em MICCAI}, pages 234--241. Springer, 2015.

\bibitem{saito2016temporal}
M.~Saito and E.~Matsumoto.
\newblock Temporal generative adversarial nets.
\newblock {\em arXiv preprint arXiv:1611.06624}, 2016.

\bibitem{salimans2016improved}
T.~Salimans, I.~Goodfellow, W.~Zaremba, V.~Cheung, A.~Radford, and X.~Chen.
\newblock Improved techniques for training gans.
\newblock In {\em Advances in Neural Information Processing Systems}, pages
  2234--2242, 2016.

\bibitem{sato2015apac}
I.~Sato, H.~Nishimura, and K.~Yokoi.
\newblock Apac: Augmented pattern classification with neural networks.
\newblock {\em arXiv preprint arXiv:1505.03229}, 2015.

\bibitem{kth2004}
C.~Schuldt, I.~Laptev, and B.~Caputo.
\newblock Recognizing human actions: a local svm approach.
\newblock In {\em ICPR}, volume~3, pages 32--36. IEEE, 2004.

\bibitem{scovanner2007}
P.~Scovanner, S.~Ali, and M.~Shah.
\newblock A 3-dimensional sift descriptor and its application to action
  recognition.
\newblock In {\em Proceedings of the 15th ACM international conference on
  Multimedia}, pages 357--360. ACM, 2007.

\bibitem{shahroudy2016ntu}
A.~Shahroudy, J.~Liu, T.-T. Ng, and G.~Wang.
\newblock Ntu rgb+ d: A large scale dataset for 3d human activity analysis.
\newblock In {\em CVPR}, pages 1010--1019, 2016.

\bibitem{simard2003best}
P.~Y. Simard, D.~Steinkraus, J.~C. Platt, et~al.
\newblock Best practices for convolutional neural networks applied to visual
  document analysis.
\newblock In {\em ICDAR}, volume~3, pages 958--962, 2003.

\bibitem{soomro2012ucf101}
K.~Soomro, A.~R. Zamir, and M.~Shah.
\newblock Ucf101: A dataset of 101 human actions classes from videos in the
  wild.
\newblock {\em arXiv preprint arXiv:1212.0402}, 2012.

\bibitem{srivastava2015unsupervised}
N.~Srivastava, E.~Mansimov, and R.~Salakhudinov.
\newblock Unsupervised learning of video representations using lstms.
\newblock In {\em International Conference on Machine Learning}, pages
  843--852, 2015.

\bibitem{szummer1996temporal}
M.~Szummer and R.~W. Picard.
\newblock Temporal texture modeling.
\newblock In {\em Image Processing, 1996. Proceedings., International
  Conference on}, volume~3, pages 823--826. IEEE, 1996.

\bibitem{tran2015learning}
D.~Tran, L.~Bourdev, R.~Fergus, L.~Torresani, and M.~Paluri.
\newblock Learning spatiotemporal features with 3d convolutional networks.
\newblock In {\em ICCV}, pages 4489--4497, 2015.

\bibitem{tulyakov2017mocogan}
S.~Tulyakov, M.-Y. Liu, X.~Yang, and J.~Kautz.
\newblock Mocogan: Decomposing motion and content for video generation.
\newblock {\em arXiv preprint arXiv:1707.04993}, 2017.

\bibitem{van2017transformation}
J.~van Amersfoort, A.~Kannan, M.~Ranzato, A.~Szlam, D.~Tran, and S.~Chintala.
\newblock Transformation-based models of video sequences.
\newblock {\em arXiv preprint arXiv:1701.08435}, 2017.

\bibitem{van2016conditional}
A.~van~den Oord, N.~Kalchbrenner, L.~Espeholt, O.~Vinyals, A.~Graves, et~al.
\newblock Conditional image generation with pixelcnn decoders.
\newblock In {\em Advances in Neural Information Processing Systems}, pages
  4790--4798, 2016.

\bibitem{villegas2017decomposing}
R.~Villegas, J.~Yang, S.~Hong, X.~Lin, and H.~Lee.
\newblock Decomposing motion and content for natural video sequence prediction.
\newblock {\em ICLR}, 1(2):7, 2017.

\bibitem{vondrick2016generating}
C.~Vondrick, H.~Pirsiavash, and A.~Torralba.
\newblock Generating videos with scene dynamics.
\newblock In {\em Advances In Neural Information Processing Systems}, pages
  613--621, 2016.

\bibitem{walker2017pose}
J.~Walker, K.~Marino, A.~Gupta, and M.~Hebert.
\newblock The pose knows: Video forecasting by generating pose futures.
\newblock {\em arXiv preprint arXiv:1705.00053}, 2017.

\bibitem{wang2011action}
H.~Wang, A.~Kl{\"a}ser, C.~Schmid, and C.-L. Liu.
\newblock Action recognition by dense trajectories.
\newblock In {\em CVPR}, pages 3169--3176. IEEE, 2011.

\bibitem{wei2000fast}
L.-Y. Wei and M.~Levoy.
\newblock Fast texture synthesis using tree-structured vector quantization.
\newblock In {\em Proceedings of the 27th annual conference on Computer
  graphics and interactive techniques}, pages 479--488. ACM
  Press/Addison-Wesley Publishing Co., 2000.

\bibitem{wong2016understanding}
S.~C. Wong, A.~Gatt, V.~Stamatescu, and M.~D. McDonnell.
\newblock Understanding data augmentation for classification: when to warp?
\newblock In {\em Digital Image Computing: Techniques and Applications (DICTA),
  2016 International Conference on}, pages 1--6. IEEE, 2016.

\bibitem{xia2012view}
L.~Xia, C.~Chen, and J.~Aggarwal.
\newblock View invariant human action recognition using histograms of 3d
  joints.
\newblock In {\em CVPRW}, pages 20--27. IEEE, 2012.

\bibitem{xue2016probabilistic}
T.~Xue, J.~Wu, K.~Bouman, and B.~Freeman.
\newblock Probabilistic modeling of future frames from a single image.
\newblock In {\em NIPS}, 2016.

\bibitem{xue2016visual}
T.~Xue, J.~Wu, K.~Bouman, and B.~Freeman.
\newblock Visual dynamics: Probabilistic future frame synthesis via cross
  convolutional networks.
\newblock In {\em Advances in Neural Information Processing Systems}, pages
  91--99, 2016.

\bibitem{yue2015beyond}
J.~Yue-Hei~Ng, M.~Hausknecht, S.~Vijayanarasimhan, O.~Vinyals, R.~Monga, and
  G.~Toderici.
\newblock Beyond short snippets: Deep networks for video classification.
\newblock In {\em Proceedings of the IEEE conference on computer vision and
  pattern recognition}, pages 4694--4702, 2015.

\bibitem{yun2012two}
K.~Yun, J.~Honorio, D.~Chattopadhyay, T.~L. Berg, and D.~Samaras.
\newblock Two-person interaction detection using body-pose features and
  multiple instance learning.
\newblock In {\em CVPRW}, pages 28--35. IEEE, 2012.

\bibitem{zhang2016stackgan}
H.~Zhang, T.~Xu, H.~Li, S.~Zhang, X.~Huang, X.~Wang, and D.~Metaxas.
\newblock Stackgan: Text to photo-realistic image synthesis with stacked
  generative adversarial networks.
\newblock {\em arXiv preprint arXiv:1612.03242}, 2016.

\bibitem{zhang2015learning}
X.~Zhang, Y.~Fu, A.~Zang, L.~Sigal, and G.~Agam.
\newblock Learning classifiers from synthetic data using a multichannel
  autoencoder.
\newblock {\em arXiv preprint arXiv:1503.03163}, 2015.

\end{thebibliography}
}

\end{document}